\pdfoutput=1
\documentclass{article}

\PassOptionsToPackage{numbers, compress}{natbib}
\usepackage[dblblindworkshop, final]{neurips_2026}

\usepackage[utf8]{inputenc}
\usepackage[T1]{fontenc}
\usepackage{hyperref}
\usepackage{url}
\usepackage{booktabs}
\usepackage{amsfonts}
\usepackage{amsmath}
\usepackage{nicefrac}
\usepackage{microtype}
\usepackage{xcolor}
\usepackage{graphicx}
\title{ToolGraph as a Q-Function for Test-Time Scaling of Tool-Use Agents}
\workshoptitle{From Pretrained Representations to Acting Agents (PTA)}

\author{%
  Ke Shen\textsuperscript{1}\qquad
  Ankita Bhaumik\textsuperscript{1}\qquad
  Li Dong\textsuperscript{1}\qquad
  Shengjie Liu\textsuperscript{1}\\
  \textbf{Shuzhen Li\textsuperscript{2}}\qquad
  \textbf{Naveed Ahmed Janvekar\textsuperscript{1}}\\[4pt]
  \textsuperscript{1}Amazon\qquad
  \textsuperscript{2}Princeton University\\
  \texttt{\{ldonga, kkshen, bhaumika, zycjlsj, njjanvek\}@amazon.com}\\
  \texttt{sl8808@princeton.edu}
}

\begin{document}

\raggedbottom

\maketitle

\begin{abstract}
Tool-use agents must chain interdependent calls, yet which outputs satisfy which
later inputs in a deployed catalog is not something a general-purpose language
model reliably infers from tool descriptions alone.  We compile that structure once
into a reusable external representation.  ToolGraph is a directed, weighted
dependency graph over a tool catalog: parameter matching and interventional probes,
which execute a candidate chain with and without its proposed predecessor, propose
edges whose evidence is combined into an edge confidence.  Prior work uses such
graphs to retrieve tools, search for tool paths, or score the next call.  We ask
instead whether the same structure can supply a \emph{multi-step} value by composing
graph paths into a goal-conditioned action value \(Q(s,a,g)\).  We study three
estimators (dynamic programming over entity sets, a GraphSAGE estimator distilled
from it, and an LLM scorer conditioned on graph evidence) plus a verbalizer that
returns action--goal feedback to the agent, which keeps final action selection.
Across two tool-use benchmarks and three backbones, we compare value ranking against
personalized-PageRank retrieval over the identical graph, and against direct and
reflective agents.  Composition pays off only when the estimator also reads the
concrete state: the LLM estimator improves on PageRank in every benchmark--backbone
setting, by \(8.3\) points on average, while the two purely structural estimators
fall below it, because a value defined over tool types cannot separate a correct tool
called with a wrong argument.  Graph construction is a one-time cost; we report it
separately rather than assume it amortized away (at our scale it is comparable to
request-time cost) and sweep four knobs controlling how much graph evidence to expose
against success and tokens.
\end{abstract}

\section{Introduction}
\label{sec:intro}

Language-model agents extend text generation into interaction by invoking tools and
incorporating their observations~\citep{yao2023react}.  The difficulty is not usually
calling a single tool correctly but \emph{chaining}: long-horizon tasks require
discovering which tool's output supplies a later tool's input, in what order, and under
which preconditions.  That dependency structure is a property of the deployed catalog
rather than of language in general, and it is only weakly recoverable from tool
descriptions read one at a time --- against an expert-annotated, parameter-level API
graph, models reading documentation alone classify dependency relations at only
42.3--70.0\% accuracy~\citep{lee2026innout} --- which is why large-catalog methods add
explicit tool representations, retrieval, or tool-use
supervision~\citep{hao2023toolkengpt,qin2024toolllm}.  The problem is sharpest in
enterprise deployments, where catalogs are large and workflows organization-specific,
but it is already measurable on public long-horizon benchmarks, which is where we study
it.

ToolGraph externalizes this structure: a node is a tool and its interface, a directed
edge records that one tool can supply information another needs.  Coupling such a graph
to a frozen language model is established
practice~\citep{chen2026gtool,patel2026dtdr,lee2026innout}; we take the graph and its
construction pipeline as given (Section~\ref{sec:toolgraph},
Appendix~\ref{app:construction}) and claim neither as a contribution, its
schema-derived edge proposal following \citet{liu2025bridging}.  Our question is what
else the same structure can do.  Prior work retrieves tools with such a graph, searches
it for tool paths~\citep{wu2024graphlearning}, or scores the call that immediately
follows~\citep{jia2026autotool}; none composes it into a multi-step, goal-conditioned
action value --- given the current state and an inferred goal, which candidate action
has the greatest expected \emph{downstream} utility?  That makes the comparison between
the two uses of one graph, ranked by query-conditioned personalized PageRank versus
ranked by a composed value, the central experiment of this paper.  It also separates two
kinds of inference cost: value-guided search and verifier-based reranking both spend
request-time compute per
request~\citep{yao2023tot,zhou2024lats,sareen2025rlv,cobbe2021verifiers,lightman2024verify,snell2024scaling,yang2026tooltree},
whereas graph construction is a one-time, catalog-level investment, so we report the two
separately rather than folding them together.

\begin{figure}[t]
  \centering
  \includegraphics[width=0.92\linewidth]{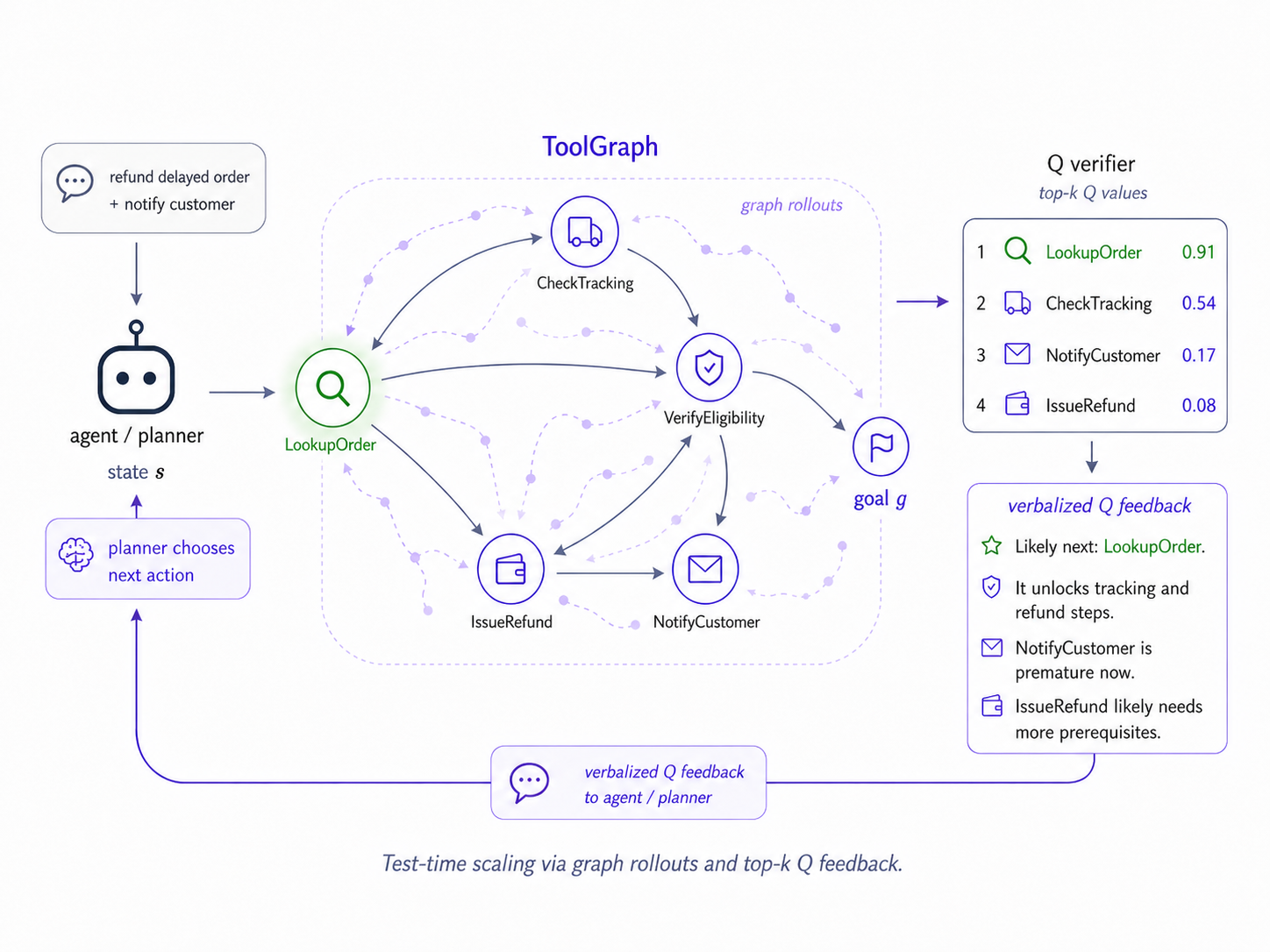}
  \caption{ToolGraph externalizes domain-specific tool dependencies and uses
  graph-derived Q-values to score candidate actions.  The top-\(k\) values
  are verbalized for the original agent, which retains final action
  selection.}
  \label{fig:overview}
\end{figure}

We study three ways to derive Q-values: analytic dynamic programming, an inductive graph
estimator, and an LLM estimator that also sees the full interaction context.  A separate
verbalizer presents action--goal values to the original agent without forcing numerical
argmax selection.  Figure~\ref{fig:overview} traces one decision: the compiled graph is
queried for paths from the entities the agent currently holds toward candidate goal
tools; each candidate action is scored by one of the estimators of
Section~\ref{sec:q-estimators}; the top-\(k\) action--goal values and their supporting
edge evidence are verbalized (Section~\ref{sec:q-verbalization}); and the planner then
chooses its next action.  Our contributions are: (i) recasting a compiled
tool-dependency graph from a retrieval structure into a source of \emph{multi-step},
goal-conditioned action values, with the controlled retrieval-versus-value comparison
over identical structure that isolates the difference; (ii) three Q-estimators spanning
auditable, learned, and semantic regimes; and (iii) a four-knob study of how much
graph-derived \emph{value evidence} to expose --- subgraph radius, edge-confidence
threshold, per-node annotation depth, and number of scored candidates --- against both
task success and token cost.

The comparison returns a qualified answer, which we state up front: composing the
graph's edges into a multi-step value beats ranking the same edges by relevance only
when the estimator can also read the concrete state.  The LLM estimator improves on
personalized-PageRank ranking of the identical graph in all six benchmark--backbone
settings, by \(8.3\) points on average, while the two purely structural estimators fall
\emph{below} it, because a value defined over tool types cannot see a correct tool called
with a wrong argument.

\section{Related Work}
\label{sec:related}

\paragraph{Tool graphs and tool retrieval.}
Large tool catalogs motivate explicit structure over tools: ToolkenGPT learns per-tool
embeddings~\citep{hao2023toolkengpt}, and ToolLLM pairs a large API corpus with a
dual-encoder retriever selecting which tools enter the prompt by query--document
similarity~\citep{qin2024toolllm} --- the structure-free control against which
graph-derived scores must be read.  Structure-aware methods couple a dependency graph to
a frozen planner in three ways.  They \emph{retrieve} with it, encoding a
request-specific graph into a continuous token~\citep{chen2026gtool} or injecting a
masked, probability-annotated tool set fitted from
demonstrations~\citep{patel2026dtdr}.  They \emph{search} it:
\citet{wu2024graphlearning} cast planning as selecting a connected subgraph of a task
graph and find a trained GNN node scorer matches or beats greedy, adaptive and beam
search while using 5--10\(\times\) fewer tokens.  Or they \emph{score} the next call
from it, as AutoTool does with a one-step transition statistic accumulated from an
agent's own executions~\citep{jia2026autotool}.  Reverse Chain is the prompt-only
precedent for our goal formulation --- pick the API that satisfies the request, then
recursively select type-matched predecessors for its unfilled arguments --- but its
backward search happens inside one generation and leaves no reusable
structure~\citep{zhang2024reversechain}, while In-N-Out supplies an expert-annotated,
parameter-level API graph as a one-hop oracle, leaving graph traversal for planning to
future work~\citep{lee2026innout}.  Two differences separate us from all of these: their
structure is mined from invocation histories, demonstrations, or expert labels, whereas
ours needs no task data; and their graph-derived quantity is a relevance score or a
one-step statistic, not a multi-step value with a goal-conditioned terminal boundary.
Our retrieval arm ranks the same graph by personalized PageRank, and it is the baseline
the value-based estimators must beat.

\paragraph{Value functions for language agents.}
The established route to an action value is gradient-based multi-turn RL: ArCHer trains
an utterance-level critic by temporal-difference backup on online, reward-labelled
rollouts~\citep{zhou2024archer}.  Closer to our deployment shape, three lines keep the
policy frozen.  \citet{hong2025planning} fit a goal-conditioned critic offline and, at
inference, verbalize its reachability estimates so a frontier model can refine its own
plan without environment search --- the nearest published neighbour to ``compiled value,
verbalized to a frozen planner.''  Self-Taught Lookahead reaches a value without reward
labels by fine-tuning a small value model to verbalize one lookahead
step~\citep{mendes2025stl}, and Reflexion is the weight-free precursor, converting
realized outcomes into verbal memory~\citep{shinn2023reflexion}.  What separates us is
not weight access --- several leave the policy untouched --- but \emph{supervision}: each
fits its value on task-specific trajectories, reward, or environment feedback, whereas
ours reads none of the three.  What we give up is expressiveness: our state abstraction
is a set of entity names, not a learned representation of the trajectory --- and
Section~\ref{sec:results} shows that this is what limits the structural estimators.

\paragraph{Verifiers, process rewards, and test-time compute.}
Sampling several candidates and selecting with a verifier is a standard way to spend
inference compute~\citep{cobbe2021verifiers}, refined by step-level process reward
models~\citep{lightman2024verify}, generative verifiers~\citep{zhang2024generative},
compute-optimal allocation studies~\citep{snell2024scaling}, and methods that reunify
reasoners with verifiers~\citep{sareen2025rlv}.  In the agentic setting,
\citet{xia2025agentrm} train a process reward model on state values extracted from search
trees and reuse one checkpoint to steer an unmodified policy; notably, an untrained judge of
the same size scores slightly \emph{below} greedy decoding there, which is why a no-graph
pole is a necessary control rather than a formality.  Our \(Q_{\mathrm{llm}}\) estimator
sits in this family, differing in that its prompt carries graph-derived evidence and a
structural prior rather than the interaction alone.  We include no trained agentic reward
model, because every published one requires the reward-labelled rollouts our setting assumes
are unavailable, and no search-based
arm~\citep{yao2023tot,zhou2024lats,yang2026tooltree}, because branching a tool environment
requires resettable state and our pilot search runs did not complete in budget.

\section{Method}
\label{sec:method}

\subsection{Problem Formulation}
\label{sec:setup}

Let \(\mathcal{T}\) be the deployed tool catalog.  At a decision point, state
\(s\) contains the user request, conversation, previous tool calls, and
observed results.  The agent considers actions
\(a\in\mathcal{A}(s)\), where each action is a tool call with arguments and
corresponds to a ToolGraph node.  For goal \(g\), \(Q(s,a,g)\) estimates the
expected future task value after taking \(a\) and continuing from the
resulting state.

ToolGraph is constructed for the deployed catalog before serving requests and
without updating the base model, injecting tool interfaces, domain relationships,
and probe evidence that need not reside in the model parameters.  During a request
it supplies features and values for the candidate actions; it is not a separate
planner, and the original agent remains responsible for its next action.

\subsection{Constructing the ToolGraph}
\label{sec:toolgraph}

ToolGraph is a directed, weighted graph over tools.  A node stores a tool's
description and input/output schemas; an edge \(A\!\rightarrow\!B\) states
that running \(A\) may supply information needed by \(B\).  We store
\[
\begin{aligned}
  v_a ={}&(\mathrm{name}_a,\mathrm{class}_a,\mathrm{domain}_a,
            \mathrm{description}_a,\mathsf{In}_a,\mathsf{Out}_a),\\
  \epsilon_{ab} ={}&(a,b,w_{ab},
            \mathsf{Match}_{ab},\mathsf{Evidence}_{ab}).
\end{aligned}
\]
\(\mathsf{In}_a,\mathsf{Out}_a\) contain field names, types, and descriptions,
and \(\mathsf{Match}_{ab}\) identifies output--input correspondences.

Two proposers run independently over tool pairs and their evidence is merged
per pair.  \emph{Parameter matching} proposes an edge when a field in
\(\mathsf{Out}_a\) satisfies a required field in \(\mathsf{In}_b\); it is cheap
but applies only to tools with well-typed declared interfaces.  An
\emph{interventional probe} treats each tool as a black-box service: a language
model derives a candidate dependency and synthesizes a probe query for it, the
query is then \emph{executed} under the composition \(A\!\rightarrow\!B\) and
under a baseline that omits \(A\), and an evaluator compares the two outcomes to
score how much \(A\) contributed.  This second route is what covers tools whose
output semantics are not declared in a schema and must be established
empirically.  Discovering edges by execution has a precedent in ToolExpNet, whose
LLM-synthesized trial-and-error episodes delete edges that do not
hold~\citep{zhang2025toolexpnet}; there the execution signal resolves a binary
keep-or-remove decision, here it is a with-versus-without-predecessor contrast
yielding the graded confidence the estimators consume.  Each retained piece
of evidence contributes a confidence, and
\(w_{ab}\) is their combination; \(\mathsf{Evidence}_{ab}\) stores the probe
query, the chain and baseline signals, and construction provenance.  Edges with
\(w_{ab}\) below a per-benchmark threshold \(\theta_{\mathrm{prune}}\) are
discarded.  Appendix~\ref{app:construction} gives the fuller walkthrough,
Table~\ref{tab:graph-stats} reports the resulting graphs, and
Table~\ref{tab:graph-cost} reports what construction cost.

Two properties of the artifact matter downstream.  First, the graph records which
outputs can satisfy which later inputs and which orderings have empirical support
--- exactly the chaining structure that is hard to read off tool descriptions
individually --- but it does \emph{not} contain complete task trajectories.  It
supplies reusable \emph{local} transition evidence, and it is the Q-estimators of
Section~\ref{sec:q-estimators} that compose that evidence across multi-tool paths.
Second, the evidence is catalog-level: edges are statements about tools, not about
particular argument values, which bounds what any graph-derived value can
discriminate --- the bound that turns out to decide our main result
(Section~\ref{sec:results}).  At decision time we retrieve paths from currently
available information toward candidate goal tools and expose their inputs, outputs,
edge confidence, and prior call outcomes.

\subsection{Graph-Derived Q Estimators}
\label{sec:q-estimators}

Graph construction provides domain-specific transition evidence; the
remaining problem is to translate it into the value of acting now.
\(Q_{\mathrm{dp}}\) explicitly composes graph paths,
\(Q_{\mathrm{emb}}\) learns transferable graph representations, and
\(Q_{\mathrm{llm}}\) combines graph evidence with the full semantic context.

\paragraph{An entity-set relaxation.}
An \emph{entity} is a named datum passed between tools, such as
\texttt{order\_id}.  Let \(E(s)\) be the entities available in state \(s\), and
let \(I_a,O_a\) be action \(a\)'s required and produced entities.  Action \(a\)
is executable when \(I_a\subseteq E(s)\).  Writing \(e=E(s)\), a successful call
yields \(e^a=e\cup O_a\) and a failed call leaves \(e\) unchanged.  Because
effects only add entities, the reachable state space is a finite lattice: this
is a probabilistic \emph{delete-free} (monotone) relaxation of the tool-use
problem, and the backup below is a probabilistic analogue of the
relaxed-reachability heuristics of classical
planning~\citep{bonet2001planning,hoffmann2001ff}.  We adopt it because it turns
valuation into a small fixed-point computation instead of a search over concrete
states, and we claim no novelty for the relaxation itself.  What is new here is that
the transition model is compiled from tool interfaces by a language model with no
trajectories and no reward labels, that goals are tools rather than declared atoms,
and that the resulting values are consumed by a frozen planner as injected evidence.

\paragraph{Goals, terminal values, and termination.}
We take \(g\) to be a \emph{goal tool}: a node whose becoming executable would
satisfy the request.  Selecting a goal API first and then working backwards over
its unfilled arguments is the strategy of Reverse
Chain~\citep{zhang2024reversechain}; we solve the same backward reachability
problem once over a persistent, weighted graph instead of inside each
generation.  Candidate goal tools \(\mathcal{G}(s)\) are inferred from
the request and the graph (Section~\ref{sec:q-verbalization}); when inference
returns none, the estimator abstains and returns a neutral score rather than a
ranking.  A state is terminal when some \(g\in\mathcal{G}(s)\) is executable in
it, i.e.\ \(I_g\subseteq e\).  Terminal states take value \(V_g=+1\) and are not
expanded further; a non-terminal state with no executable action is a dead end
with \(V_g=-1\).  Values therefore lie in \([-1,1]\), and goal attainment enters
only through the terminal boundary: the immediate reward is identically zero, so
Eq.~\eqref{eq:q-dp} carries no separate reward term.

\paragraph{Analytic dynamic programming (\(Q_{\mathrm{dp}}\)).}
For a non-terminal \(e\),
\begin{equation}
  \bar Q_g(e,a) =
  \gamma\left[
    p_a V_g(e^a)+(1-p_a)V_g(e)
  \right],
  \qquad
  V_g(e)=\max_{a:\,I_a\subseteq e} q_g(e,a),
  \label{eq:q-dp}
\end{equation}
with \(V_g(e)=+1\) on terminal states and \(-1\) when no action is executable.
The failure branch \((1-p_a)V_g(e)\) is a self-loop in the \emph{value
recursion}, not a licence to retry an action indefinitely in an episode.  We obtain a
fixed point of Eq.~\eqref{eq:q-dp} by value iteration over the reachable lattice; with
\(\gamma<1\) the unshrunk backup is a contraction, while the shrinkage of
Eq.~\eqref{eq:q-dp-penalty} can inflate the modulus to \(\gamma(1+\kappa)\), so we
verify convergence numerically at tolerance \(10^{-6}\) rather than assert uniqueness
from \(\gamma<1\) alone.  We
enumerate that lattice by breadth-first closure from the entity set the agent
currently holds, stopping at terminal states and under an explicit cap on the
number of enumerated states; Appendix~\ref{app:impl} reports that cap and the
solver's convergence settings.  Because a required
field that no tool produces can only come from the conversation, reachability is
computed over tool-producible entities only, and conversation-supplied arguments
are checked instead at the candidate level below.

\paragraph{Confidence shrinkage.}
Graph edges carry a confidence, and we damp values that rest on weak or
inconsistent edges:
\begin{equation}
  q_g(e,a)=
  \bar Q_g(e,a)-\kappa(1-c_a)\bigl|\bar Q_g(e,a)\bigr|,
  \qquad \kappa\in[0,1].
  \label{eq:q-dp-penalty}
\end{equation}
This is not a lower confidence bound.  Because the penalty scales \(|\bar Q_g|\) but
is subtracted, it damps optimistic values toward zero and \emph{deepens} pessimistic
ones: for \(\bar Q_g>0\) the factor is \(1-\kappa(1-c_a)\), for \(\bar Q_g<0\) it is
\(1+\kappa(1-c_a)\).  It is therefore a magnitude penalty on dispersion in the
supporting evidence, not a shrinkage toward neutral, and we clamp to \([-1,1]\) before
the affine map below.  Because \(V_g\) feeds back into Eq.~\eqref{eq:q-dp}, the factor
applies once per backup step, so a value supported by a \(d\)-step chain is penalized
more than a one-step value; this penalizes long dependency chains.  We fix \(\kappa=0.3\) rather than tuning it, and
we do not sweep it; since the structural estimators lose to relevance ranking on the
graph regardless (Section~\ref{sec:results}), a \(\kappa\) sweep is unlikely to be
where their gap lies.  We also note that \(p_a\) and \(c_a\) are not independent
evidence channels: both are aggregations of the \emph{same} incoming edge
weights for \(a\) --- \(p_a\) their maximum, \(c_a\) their mean --- so
Eq.~\eqref{eq:q-dp-penalty} acts as a dispersion penalty over that evidence
rather than as a second measurement.  Tools with no incoming edges receive a
neutral value for both.  Runtime evidence enters as inputs, not structure: a
tool that has failed in this episode and never succeeded is assigned
\(p_a=0\).

\paragraph{From values to candidate scores.}
Candidates are proposed actions, which may be tool calls or a message to the
user.  We set \(Q_{\mathrm{dp}}(s,a,g)=q_g(E(s),a)\) and map \([-1,1]\) onto
\([0,1]\) monotonically.  Three cases are handled explicitly.  A candidate whose
tool is absent from the graph, whose required fields are neither available nor
supplied by a conversation-grounded argument, or which has been marked a dead
end, is scored strictly worst; this admissibility filtering is a by-product of the
relaxation rather than an objective.  A message-only candidate is scored
\(\gamma V_g(E(s))\), the value of leaving the entity state unchanged, which places
it between goal-directed and inadmissible calls --- a modelling assumption, since a
catalog-level graph cannot assess conversational value.  A candidate proposing
several calls
composes them sequentially, with \(p\) the product of the per-tool values and
\(c\) their minimum.  This estimator is fast, auditable, and makes no model
calls, but it cannot distinguish two candidates that call the same tool with
different argument values, nor two states with the same fields and different
histories.

\paragraph{Inductive graph estimator (\(Q_{\mathrm{emb}}\)).}
We learn GraphSAGE~\citep{hamilton2017inductive} over schema embeddings,
probe scores, degree statistics, and input/output counts.  Let \(z_a\) embed
the action, \(z_s\) pool successful tools, and \(z_g\) pool inferred goals:
\begin{equation}
  Q_{\mathrm{emb}}(s,a,g)
  = \sigma\!\left(\operatorname{MLP}
    [z_a\,\Vert\,z_s\,\Vert\,z_g]\right).
  \label{eq:q-emb}
\end{equation}
We distill normalized \(Q_{\mathrm{dp}}\) targets over reachable states.  Training a
graph network over a tool graph to score nodes for a frozen planner is not
new---\citet{wu2024graphlearning} compare exactly this family, including
GraphSAGE---and the difference here is the target: our regression labels are
distilled \(Q_{\mathrm{dp}}\) values, which require no task annotation, rather than a
ranking loss on gold step--tool pairs.  The motivation is amortization rather than
generalization to new catalogs: once trained, \(Q_{\mathrm{emb}}\) scores a candidate
with a single forward pass and no lattice enumeration, whereas \(Q_{\mathrm{dp}}\)
re-solves per decision and per goal.  Neighbourhood aggregation additionally lets it
assign a value to a tool the relaxation leaves unvalued, provided the tool is
connected.  We do not test transfer to unseen catalogs.

\paragraph{Language-model estimator (\(Q_{\mathrm{llm}}\)).}
The LLM sees the full state, candidate arguments, goal, policy, alternatives,
verbalized graph evidence, and the \(Q_{\mathrm{dp}}\) value as a prior.  It
returns a 0--10 score and rationale; several scorer samples are averaged and
normalized to \([0,1]\).  This lets the estimator account for concrete entity
values, conversation state, and policies that catalog-level graph structure does
not represent, at higher request-time cost.  Note that the three estimators are
nested rather than independent --- \(Q_{\mathrm{emb}}\) is distilled from
\(Q_{\mathrm{dp}}\), and \(Q_{\mathrm{llm}}\) is conditioned on it --- so the
comparison among them is a comparison of how much each adds over the one it
contains, not of three independent methods.  The flat retrieval control of
Section~\ref{sec:exp-setup} is our no-graph pole; we discuss there what that
control does and does not establish about a graph-free scorer.

\subsection{Top-k Retrieval and Natural-Language Q Feedback}
\label{sec:q-verbalization}

Let \(\mathcal{G}(s)\) be the candidate goals inferred from the request and
graph.  We cheaply recompute \(Q_{\mathrm{dp}}\) for each goal to form
\begin{equation}
  M[a,g]=Q_{\mathrm{dp}}(s,a,g),
  \qquad a\in\mathcal{A}(s),\;g\in\mathcal{G}(s).
  \label{eq:q-matrix}
\end{equation}
A verbalizer converts each row and its graph evidence into action--goal
feedback for the agent.  It neither selects the argmax nor estimates a new
value; unlike \(Q_{\mathrm{llm}}\), it only translates existing values.  Its output
is feedback plus a recommended action, which the planner may disregard, and it costs
one model call per decision --- included in the token accounting of
Section~\ref{sec:ablations}.

Three distinct budgets appear below and we fix the notation here.  The
\emph{injection width} \(k\) is the number of action--goal rows placed in the
prompt, each carried with its own retrieved subgraph, whose size is controlled
separately by hop depth \(h\) and edge threshold \(\theta\); the \emph{candidate
count} \(n\) is the number of actions the planner proposes per decision for the
estimator to rank; and the number of \(Q_{\mathrm{llm}}\) scorer samples averaged
per candidate controls scoring fidelity.  These give modest task-time compute
controls without requiring online environment search.

\section{Experiments}
\label{sec:experiments}

\subsection{Setup}
\label{sec:exp-setup}

\paragraph{Evaluation.}
We use ToolComp~\citep{nath2025toolcomp} and
\(\tau^2\)-Bench~\citep{barres2025tau2bench} with Ministral 3 8B, Gemma 4
31B, and DeepSeek V3.2 685B.  Baselines are Native, ReAct
\citep{yao2023react}, and Reflexion~\citep{shinn2023reflexion}, covering direct
action generation, reasoning with observations, and reflection.

Two controls isolate what the graph contributes.  \textbf{Flat top-\(k\)
retrieval} ranks tools by embedding similarity between the request and each
tool's description and injects the top \(k\) into the planner's context, with no
graph, no propagation, and no scorer --- the structure-free counterpart of the
retrieval interface used by dual-encoder tool
retrievers~\citep{qin2024toolllm}.  It is obtained from the same code path by
removing graph propagation, so the embedding, \(k\), and prompt template are
identical to the ToolGraph arms; the difference is only whether dependency
structure informs the ranking.  Its purpose is to separate the \emph{interface} from
the \emph{structure}, so that any gain the ToolGraph arms show over it is attributable
to the graph rather than to \(k\) tools being in the context.  It is also our stand-in
for deleting the graph from \(Q_{\mathrm{llm}}\), on the untested assumption that a
scorer with no graph evidence adds little over similarity ranking; if that is wrong,
the flat arm understates the no-graph condition.  \textbf{TG--PPR} then ranks the same
catalog by query-conditioned personalized PageRank over the weighted graph.  Every
ToolGraph variant shares the graph, the candidate pool, the verbalizer, and the
top-\(k\) interface; only the ranking changes, although \(Q_{\mathrm{llm}}\) also spends
request-time scorer calls that the other arms do not.  Flat versus TG--PPR therefore isolates
dependency structure from plain relevance, and TG--PPR versus the three estimators
isolates multi-step value composition from local graph relevance.  Each ToolGraph is
constructed once per benchmark and reused across tasks and backbones.

We select the injection width \(k=8\) on a stratified development split (20\% of
non-test tasks) and preserve official test partitions.  Every arm draws \(n=4\)
candidate actions per decision, so the estimators rank identical candidate sets, and
Table~\ref{tab:sweeps} sweeps \(n\).  Success is the benchmark-native binary reward on
the \(\tau^2\) test split of 100 tasks (airline 20, retail 40, telecom 40) and the
ToolComp evaluation split of 214 tasks; construction cost is reported separately from
request-time tokens.  Three models are held fixed while only the planner
varies: the scorer and verbalizer (Gemma 4 31B), the \(\tau^2\) user simulator
(Sonnet 4.5), and the ToolComp answer judge.

\subsection{Results}
\label{sec:results}

\begin{table}[t]
  \centering
  \caption{Task success rate (\%).  Flat top-\(k\) retrieval and the four ToolGraph
  rows share the same \(k=8\) injection interface and prompt template; they differ
  only in how the injected tools are ranked, and only the ToolGraph rows use the
  graph.  M, G, and D denote Ministral 3 8B, Gemma 4 31B, and DeepSeek V3.2 685B.
  Bold marks the best value in each column.  Verifier:
  Gemma 4 31B for every scored arm; ToolComp answers are additionally graded by the
  benchmark's own answer judge, held fixed across arms.}
  \label{tab:main}
  \small
  \setlength{\tabcolsep}{4pt}
  \begin{tabular}{lcccccccr}
    \toprule
    & \multicolumn{3}{c}{ToolComp}
    & \multicolumn{3}{c}{\(\tau^2\)-Bench}
    & \\
    \cmidrule(lr){2-4}\cmidrule(lr){5-7}
    Method & M & G & D & M & G & D & Mean \\
    \midrule
    Native                            & 56.9 & 72.6 & 73.1 & 40.0 & 67.0 & 72.0 & 63.6 \\
    ReAct                             & 46.7 & 65.0 & 62.4 & 38.0 & 52.0 & 63.0 & 54.5 \\
    Reflexion                         & 50.8 & 69.5 & 72.6 & 49.0 & 69.0 & 75.0 & 64.3 \\
    \midrule
    Flat top-\(k\) retrieval          & 53.8 & 71.6 & 72.1 & 34.0 & 62.0 & 72.0 & 60.9 \\
    \midrule
    TG top-\(k\) PPR                  & 54.8 & 74.1 & 69.5 & 37.0 & 66.5 & 74.0 & 62.7 \\
    TG top-\(k\) \(Q_{\mathrm{dp}}\)  & 55.8 & 71.6 & 73.6 & 38.0 & 54.0 & 66.6 & 59.9 \\
    TG top-\(k\) \(Q_{\mathrm{emb}}\) & 57.6 & 69.5 & 68.4 & 36.0 & 50.4 & 66.0 & 58.0 \\
    TG top-\(k\) \(Q_{\mathrm{llm}}\) & \textbf{62.0} & \textbf{79.7} & \textbf{78.0}
      & \textbf{53.0} & \textbf{73.0} & \textbf{80.0} & \textbf{71.0} \\
    \bottomrule
  \end{tabular}
\end{table}

Table~\ref{tab:main} is laid out so that three successive differences each isolate one
step.

\paragraph{The interface costs success; dependency structure recovers most of it.}
Native to flat top-\(k\) retrieval isolates the \emph{interface}: \(k\) tool
descriptions in the context, ranked by request--description similarity, with no
dependency structure.  It \emph{loses} in five of six settings (\(-3.1/-1.0/-1.0\) on
ToolComp, \(-6.0/-5.0/0.0\) on \(\tau^2\); Mean \(63.6\to60.9\)).  Since every arm
already sees the full domain catalog (Appendix~\ref{app:impl}), the injected block adds
no callable tool; it competes for attention, and a relevance ranking that ignores
dependencies sometimes promotes a tool that cannot yet be called.  Flat retrieval to
TG--PPR then holds the interface, embedding, and \(k\) fixed and changes only whether
dependency structure informs the ranking: it helps in five of six settings
(\(+1.0/+2.5/-2.6\), \(+3.0/+4.5/+2.0\); Mean \(60.9\to62.7\)), so the structure is
real and readable by a frozen planner --- though it does not fully repay the
interface, TG--PPR staying \(0.9\) points below Native on the Mean.  Together these set
the bar the value arms must clear: not whether graph evidence beats nothing, but
whether it beats the same interface filled by relevance.

\paragraph{Composing edges into a value helps only when the estimator also reads the
state.}
TG--PPR to the three estimators is the central contrast, and the estimators see no
information the retrieval arm lacks --- only a different way of combining it.
\(Q_{\mathrm{dp}}\) splits its six cells against TG--PPR three--three
(\(+1.0/-2.5/+4.1\), \(+1.0/-12.5/-7.4\)) and loses \(2.8\) points on the Mean only
because its two losses are far larger than its three wins; \(Q_{\mathrm{emb}}\) wins one
cell of six and loses \(4.7\) on the Mean.  \(Q_{\mathrm{llm}}\), which sees the same
graph evidence plus the concrete state, beats TG--PPR in \emph{every} setting
(\(+7.2/+5.6/+8.5\) on ToolComp, \(+16.0/+6.5/+6.0\) on \(\tau^2\); Mean
\(62.7\to71.0\)) and is the best arm in every column and on the Mean, ahead of the
strongest baseline, Reflexion, by \(6.7\) points.

Two effects are at work and should not be conflated.  Argument blindness is
\emph{shared} by TG--PPR and the structural estimators --- both rank tool types --- so
it cannot explain why composition loses to relevance; it explains the ceiling both hit,
and why only the state-reading scorer clears it.  What separates them is the
relaxation's \emph{bias}: a monotone reachability value rewards actions that unlock
many downstream types, which is not relevance to this request, and multi-step
propagation compounds that preference along a path rather than damping it.  Composition
therefore does not merely fail to add signal, it moves the ranking --- so
\(Q_{\mathrm{dp}}\) earns its place as the scorer's prior and admissibility filter, not
as a ranker, and \(Q_{\mathrm{emb}}\) inherits the bias without the auditability.  One
caveat: cells are point estimates without intervals, so we do not order small
differences.

\subsection{Ablations and Analysis}
\label{sec:ablations}

Taking \(Q_{\mathrm{llm}}\) as \(Q_{\mathrm{best}}\) on the strength of
Table~\ref{tab:main}, we vary the \emph{evidence budget} on \(\tau^2\) along four
knobs.  Three control how much of the graph the scorer sees per decision: the
neighbourhood radius \(h\) of the subgraph extracted around the candidate and goal
tools, the edge-weight threshold \(\theta\) below which edges are pruned before
verbalization, and the per-node annotation depth --- edges only (\(\mathsf{N0}\)),
plus required/produced fields (\(\mathsf{N1}\)), structural features
(\(\mathsf{N2}\)), or runtime call history (\(\mathsf{N3}\)).  The fourth, \(n\), is
the number of candidate actions sampled per decision for the scorer to rank;
\(n{=}1\) leaves the verbalizer's feedback as the only graph signal.  The first two
knobs act indirectly, so we report the resulting node and edge counts.

Rather than cross the knobs, we run four one-factor sweeps through an \emph{anchor}
configuration --- fourteen configurations instead of the \(4\times6\times4\times3\)
cross.  The sweeps run on the evaluation split, so the anchor is a reported operating
point, not one held out from these numbers.  This answers what the scaling claim needs
--- which knob returns more success per token --- but cannot detect interaction, and it
characterizes each knob only near the anchor and only for \(Q_{\mathrm{llm}}\), since
the structural estimators read the graph directly rather than through a prompt.  Cost is
request-time tokens per task, \(\sum_i t_i\) over candidate generation, scoring and
verbalization; the one-time construction term \(C\), amortized as \(C/N\) over \(N\)
reusing tasks, is kept separate in Table~\ref{tab:graph-cost} since it shifts every
ToolGraph arm equally.  We keep it separate rather than declaring it amortized away
because at our scale it is not: over the \(100\) evaluated \(\tau^2\) tasks \(C/N\) is
\(88\)K tokens, comparable to the anchor's own \(155\)K of request-time cost.  It falls
as \(1/N\), so the argument is about deployments serving many more requests than a test
split, not about these cells.

\begin{table}[t]
  \centering
  \caption{Evidence-budget sweeps for $Q_{\mathrm{llm}}$ on $\tau^2$-Bench.  Each block
  varies \emph{one} knob and holds the rest at the anchor (\dag: $h{=}1$,
  $\theta{=}0.0$, $\mathsf{N0}$, $n{=}4$).  Success is the macro average of task pass
  rate (\%) over three planners and the three $\tau^2$ domains; Tok.\,K is an
  \emph{estimated} request-time cost in thousands of tokens per task (planner, scorer
  and verbalizer at ${\approx}5.5$ decisions/task), computed from measured prompt sizes
  and per-mechanism call counts rather than per-call accounting.  Blocks 1--3 and block
  4 are separate run series, so the nominally identical anchor reads $65.8$ in the first
  three and $66.4$ in the fourth --- a $0.6$-point gap \emph{on the macro average};
  the same replicate pair differs by up to $2.5$ points within a single planner column,
  which is the scale at which per-backbone differences should be read.  Sweeps are run
  on the evaluation split.  Verifier: Gemma 4 31B; verbalizer enabled throughout.}
  \label{tab:sweeps}
  \small
  \setlength{\tabcolsep}{3.5pt}
  \begin{tabular}{@{}lccc@{\hspace{14pt}}lccc@{}}
    \toprule
    & nodes/edges & Succ. & Tok.\,K & & nodes/edges & Succ. & Tok.\,K \\
    \midrule
    \multicolumn{4}{@{}l}{\textbf{1 --- radius $h$}}
      & \multicolumn{4}{l@{}}{\textbf{2 --- edge threshold $\theta$}} \\
    $h{=}0$                & 7.7/7.0     & 64.4 & 136.9 & $\theta{=}0.0$\,\dag & 24.1/159.5 & \textbf{65.8} & 155.2 \\
    $h{=}1$\,\dag          & 24.1/159.5  & \textbf{65.8} & 155.2 & $\theta{=}0.5$ & 24.1/149.4 & 65.3 & 153.9 \\
    $h{=}2$                & 24.3/160.3  & 64.5 & 155.2 & $\theta{=}0.6$ & 24.1/138.4 & 63.6 & 152.7 \\
    $h{=}3$                & 24.3/160.3  & 63.6 & 155.2 & $\theta{=}0.7$ & 24.1/131.4 & 63.0 & 152.0 \\
                           &             &      &       & $\theta{=}0.8$ & 24.1/125.0 & 63.6 & 151.3 \\
                           &             &      &       & $\theta{=}0.9$ & 24.1/\phantom{0}73.7 & 63.3 & 145.7 \\
    \midrule
    \multicolumn{4}{@{}l}{\textbf{3 --- node annotation depth}}
      & \multicolumn{4}{l@{}}{\textbf{4 --- sampled candidates $n$}} \\
    $\mathsf{N0}$ edges only\,\dag & 24.1/159.5 & 65.8 & 155.2 & $n{=}1$ verbalizer only & 24.1/159.5 & 61.1 & \phantom{0}67.7 \\
    $\mathsf{N1}$ +req./prod.      & 24.1/159.5 & \textbf{66.4} & 158.5 & $n{=}4$\,\dag & 24.1/159.5 & \textbf{66.4} & 155.2 \\
    $\mathsf{N2}$ +structural      & 24.1/159.5 & 64.2 & 161.1 & $n{=}8$ & 24.1/159.5 & 64.7 & 271.8 \\
    $\mathsf{N3}$ +runtime         & 24.1/159.5 & 63.3 & 163.7 & & & & \\
    \bottomrule
  \end{tabular}
\end{table}

Every knob buys evidence at a token price, so each block is read against its own anchor:
the change in success per additional thousand tokens is what that knob returns, and a
block in which success falls as the budget grows is one where the extra evidence buys
dilution rather than dependencies.  Three things hold of the macro average across all
four blocks; per backbone the peaks move, so these are aggregate tendencies rather than
per-model prescriptions.  First, \emph{every knob except \(\theta\) is non-monotone, with its macro
peak at or near the second level} --- \(h{=}1\), \(\mathsf{N1}\), \(n{=}4\): additional
structure first supplies missing dependencies and then crowds the prompt, and four to
six levels expose a peak without locating it precisely.  Second, \emph{the peak is also
the knee}, so the defensible operating choice is never the most expensive end:
\(h{=}2\) and \(h{=}3\) cost exactly the same \(155.2\)K as \(h{=}1\) and lose
\(1.3\) and \(2.2\) points, because the \(\tau^2\) graph is already almost covered at
one hop (\(24.1\to24.3\) nodes), so further radius buys no structure and only reorders
the prompt; and \(n{=}8\) spends \(116.6\)K more than \(n{=}4\) to lose \(1.7\) points.
Third, \emph{pruning to save tokens never pays}: raising \(\theta\) to \(0.9\) removes
over half the edges to save \(9.5\)K tokens and costs \(2.5\) points, declining almost
monotonically in between, so the weak edges carry signal rather than noise.

Ranked by marginal return, node annotation is the cheapest win --- \(\mathsf{N0}\) to
\(\mathsf{N1}\) buys \(0.6\) points for \(3.3\)K tokens by telling the scorer \emph{why}
two tools connect, not just \emph{that} they do --- ahead of radius (\(1.4\) for
\(18.3\)K) and candidate count (\(5.3\) for \(87.5\)K), the largest absolute effect.
With no candidate diversity at all the verbalizer's feedback still reaches \(61.1\),
nine tenths of the anchor's success at under half its budget.

\section{Limitations}
\label{sec:limitations}

ToolGraph supplies only structure its nodes and edges represent correctly, and two
blind spots are structural rather than incidental: the delete-free relaxation ignores
consumed resources, conditional effects, and negative interactions between
actions~\citep{hoffmann2005ignoring}, so it overvalues plans a non-monotone model
would reject; and edges are catalog-level statements about tool pairs, so no
graph-derived value separates two candidates calling the same tool with different
arguments --- the limit that decides our main result.  Probe outcomes are adjudicated
by a model against synthesized queries rather than gold answers, so an edge weight is
evidence of a dependency and not a proof of one; probing is harder for stateful or
irreversible APIs, and amortization assumes a stable catalog and enough reuse.  Our
benchmarks are public, so we test only the narrower claim that composing catalog
dependencies into a value beats using the same graph for retrieval, and our retrieval
pole is a same-graph ablation rather than the state of the art~\citep{patel2026dtdr}.
Holding one scorer model fixed across planner backbones lets the LLM-scored arm
benefit from a scorer stronger than the planner it advises, a confound the structural
estimators do not have.  Finally, cells are single-seed point estimates without
confidence intervals, the sweeps cannot detect interaction between knobs, we do not
test \(Q_{\mathrm{emb}}\) transfer to unseen catalogs, and we cannot release code,
graph artifacts, or trajectories.

\section{Conclusion}
\label{sec:conclusion}

We asked whether a compiled tool-dependency graph, which prior work retrieves from,
searches, or reads a single next call out of, can instead be composed into a multi-step
goal-conditioned action value.  The answer comes back split: over one fixed graph the LLM
estimator beats same-graph personalized-PageRank retrieval in every one of six settings, by
\(8.3\) points on the mean, while the auditable fixed-point solve and its distilled variant
do not.
The useful reading of a compiled dependency graph is therefore neither ``retrieve from
it'' nor ``rank by its value'' but the composition of the two: the structural solve
supplies a cheap, auditable prior and admissibility filter, and a scorer that can still
see the arguments turns it into a decision.  That also says where the next gain is ---
lifting the entity-set relaxation toward argument-level state, rather than enlarging the
graph.

\bibliographystyle{plainnat}
\bibliography{refs}

\appendix

\section{ToolGraph Construction}
\label{app:construction}

This appendix summarizes how the graphs used in our experiments were built.  It
is included for self-containment; the pipeline is not a contribution of this
paper, and we describe it only to the depth needed to make the Q-estimators of
Section~\ref{sec:q-estimators} well defined.

\paragraph{Nodes.}
Every tool is normalized into a common record regardless of its source package:
identifier, a coarse class distinguishing \emph{atomic} tools with declared,
well-typed interfaces from \emph{meta} tools whose output semantics are not
declared and must be established empirically, a domain grouping, a textual
statement of what the tool does, and the names, types, and descriptions of its
input and output fields.  A node embedding is formed from the concatenated
intent and output text and is reused as the input feature for
\(Q_{\mathrm{emb}}\).

\paragraph{Edge proposal.}
For an ordered pair \((a,b)\) we ask whether \(a\) should precede \(b\), whether
\(b\) should precede \(a\), or whether the two are independent or substitutable
and no edge should exist.  Two proposers run independently.  \emph{Parameter
matching} compares \(\mathsf{Out}_a\) against the required fields of
\(\mathsf{In}_b\) and proposes an edge on a match; it is nearly free but is
restricted to tools that declare usable types, so it does not reach meta tools.
\emph{Interventional probing} treats each tool as a black-box service.  A
language model derives a candidate dependency from one tool's interface --- what
information must be obtained before it can be called, or what task naturally
follows from its output --- checks whether the other tool could plausibly supply
that piece, and synthesizes a probe query if so.  The probe query is then
executed under the composition and under a baseline that omits the proposed
predecessor, and an evaluator compares how much of the query each outcome
resolves.  The two proposers overlap: many pairs are proposed by probing alone,
which is why probing rather than parameter matching is the primary source of
edges (Table~\ref{tab:graph-stats}).

\paragraph{Edge confidence and pruning.}
Rather than a binary decision, each edge carries a confidence in \([0,1]\).  The
capability-match score acts as a prior, the probe comparison is converted into a
likelihood, and the two are combined per piece of evidence; \(w_{ab}\) is the
mean over the retained evidence components for that pair.  An edge is
instantiated only when \(w_{ab}>\theta_{\mathrm{prune}}\).  The threshold is set
per benchmark and its values are given in Table~\ref{tab:graph-stats}; the
sparser the candidate set, the lower the threshold needed to retain a usable
graph.  The graph is built once per benchmark, frozen, and reused by every task,
backbone, and ToolGraph arm in our experiments; we perform no runtime updates to
edge weights.

\paragraph{Provenance.}
Each edge stores the probe query, the chain and baseline signals, the resulting
confidence components, and a graph version hash, so that a value produced by
\(Q_{\mathrm{dp}}\) can be traced back to the specific evidence supporting each
edge on the path.  This is what we mean by calling the structural estimator
auditable.

\paragraph{Separation from the evaluation data.}
Probe queries are synthesized from tool interfaces alone.  No benchmark task
text, reference trajectory, reward signal, or environment feedback from the
evaluation split is read at any point during construction, so edge weights
cannot encode information about the tasks the graph is later evaluated on.  This
matters because a single graph per benchmark is reused across every task and
backbone: had construction observed test tasks, that leakage would inflate every
ToolGraph arm simultaneously.

\begin{table}[t]
  \centering
  \caption{ToolGraph statistics per benchmark.  ``Param.\ only'' / ``probe only'' /
  ``both'' partition the retained edges by which proposer supplied their evidence.
  The lower block gives the three \(\tau^2\) domains that are evaluated separately,
  counting only intra-domain nodes and edges; the \(73\) domain nodes are a subset of
  the \(146\) in the catalog graph, the remainder belonging to \(\tau^2\) domains we do
  not evaluate.  In that block the last column is \emph{not}
  \(\theta_{\mathrm{prune}}\) but \(k/|\mathcal{T}|\), the fraction of the domain
  catalog exposed at \(k=8\), which shows where retrieval has little room to act.}
  \label{tab:graph-stats}
  \small
  \setlength{\tabcolsep}{4pt}
  \begin{tabular}{lrrrrrrrr}
    \toprule
    & \multicolumn{2}{c}{Nodes}
    & \multicolumn{2}{c}{Edges}
    & \multicolumn{3}{c}{Evidence source}
    & \\
    \cmidrule(lr){2-3}\cmidrule(lr){4-5}\cmidrule(lr){6-8}
    Benchmark & atomic & meta & cand. & kept
      & param.\ only & probe only & both
      & \(\theta_{\mathrm{prune}}\) \\
    \midrule
    ToolComp    &   7 &  4 &    40 &    35 &   0 &   28 &    7 & 0.10 \\
    \(\tau^2\)-Bench & 102 & 44 &  1508 &  1303 &  59 & 1041 &  203 & 0.30 \\
    \midrule
    \multicolumn{9}{l}{\emph{Evaluated \(\tau^2\) domains (intra-domain edges;
      \(k/|\mathcal{T}|\) at \(k=8\) in the last column)}} \\
    \quad airline &  \multicolumn{2}{c}{14} & \multicolumn{2}{c}{79}
      & \multicolumn{3}{c}{} & 0.57 \\
    \quad retail  &  \multicolumn{2}{c}{16} & \multicolumn{2}{c}{93}
      & \multicolumn{3}{c}{} & 0.50 \\
    \quad telecom &  \multicolumn{2}{c}{43} & \multicolumn{2}{c}{309}
      & \multicolumn{3}{c}{} & 0.19 \\
    \bottomrule
  \end{tabular}
\end{table}

\begin{table}[t]
  \centering
  \caption{Graph construction and maintenance cost, in language-model calls and
  estimated tokens, broken out by construction phase.  Construction is a one-time,
  catalog-level cost; the amortized rows divide it by the number of evaluated tasks
  that reuse the graph.  On the larger catalog, adding a tool later costs about two
  orders of magnitude less than a rebuild and deleting one is near-free, which is what
  makes the compiled artifact reusable under catalog churn; on a catalog as small as
  ToolComp's eleven tools the saving is correspondingly smaller.  Token counts are estimates, derived from
  measured prompt sizes and the max-token setting of each phase rather than from
  per-call accounting.  Construction uses a fixed closed model (Sonnet~4.5 for edge
  proposal and probe synthesis, Sonnet~4 for probe evaluation), independent of the
  planner and verifier models used at request time.}
  \label{tab:graph-cost}
  \small
  \setlength{\tabcolsep}{5pt}
  \begin{tabular}{lrr}
    \toprule
    & ToolComp & $\tau^2$-Bench \\
    \midrule
    \multicolumn{3}{l}{\textbf{Full construction (one-time)}} \\
    \midrule
    Phase~2: schema edge evaluation      & 21 & 1{,}266 \\
    Phase~3.1: blind task derivation     &  7 &   102 \\
    Phase~3.2: capability matching       & 28 & 4{,}488 \\
    Phase~3.3: information gain          & 65 & 2{,}400 \\
    \cmidrule{2-3}
    \textbf{Total LLM calls}             & \textbf{121} & \textbf{8{,}256} \\
    \quad Input tokens                   & 83K & 6.0M \\
    \quad Output tokens                  & 37K & 2.8M \\
    \quad Total tokens                   & 120K & 8.8M \\
    \midrule
    \multicolumn{3}{l}{\textbf{Incremental update}} \\
    \midrule
    LLM calls, add one tool              &  14 &   84 \\
    \quad Total tokens                   & 14K & 89K \\
    LLM calls, delete one tool           &   7 &    7 \\
    \midrule
    Evaluated tasks reusing the graph    & 214 & 100 \\
    Amortized construction calls/task    & 0.6 & 83 \\
    Amortized construction tokens/task   & 0.6K & 88K \\
    \bottomrule
  \end{tabular}
\end{table}
\section{Implementation Details}
\label{app:impl}

\paragraph{Graph construction.}
The pipeline is described in Appendix~\ref{app:construction} and the resulting
graphs are characterized in Table~\ref{tab:graph-stats}.  What matters for the
estimators is how per-edge confidences become per-action quantities.  Both
\(p_a\) and \(c_a\) are aggregations over the weights of the edges \emph{incoming}
to \(a\): \(p_a\) is their maximum and \(c_a\) their mean.  They are therefore
two summaries of one evidence array rather than two independent measurements,
which is why Eq.~\eqref{eq:q-dp-penalty} functions as a dispersion penalty; we
state this explicitly because it bounds how much additional discrimination the
penalty can provide.  A tool with no incoming edges receives the neutral value
\(0.5\) for both.  At decision time the retriever constructs a pool of at most
16 feasible or goal-relevant actions.

\paragraph{Q estimators.}
For \(Q_{\mathrm{dp}}\), we use \(\gamma=0.95\) and \(\kappa=0.3\), with a
convergence tolerance of \(10^{-6}\), a cap of \(10^{4}\) value-iteration sweeps,
and a cap of \(10^{5}\) enumerated entity-set states; the solver raises rather
than silently truncating if the cap is reached.  Terminal states take value
\(+1\) and dead ends \(-1\).  Solutions are cached per decision state and
re-solved only when a candidate's successor state falls outside the enumerated
lattice.  Because the solver raises rather than truncating, every reported cell is
one in which the cap was not reached.
\(Q_{\mathrm{emb}}\) uses two mean-aggregation GraphSAGE layers with hidden
dimension 256, 0.1 dropout, and a three-layer regression head.  Node input
combines a 1024-dimensional schema embedding with eight normalized structural
features.  We train for 200 epochs with Adam, learning rate \(10^{-3}\),
weight decay \(10^{-5}\), batch size 64, and mean-squared error against
normalized \(Q_{\mathrm{dp}}\) targets.  Training states and goal tools are
enumerated from development-split states only, so that no test-split state
contributes a distillation target.

For \(Q_{\mathrm{llm}}\), each candidate prompt contains the current request,
conversation and tool history, candidate arguments, inferred goals, policy
text, feasible alternatives, graph paths, edge evidence, and
\(Q_{\mathrm{dp}}\) prior.  The scorer returns JSON with an integer in
\([0,10]\) and a short rationale.  We average five samples at temperature
1.0 and normalize by 10.  The same verifier model and prompt template are
used for all planner backbones.

\paragraph{Agent and baseline configuration.}
The ToolGraph methods place the top-\(k\) actions, scores, goals, and
verbalized evidence in the planner prompt; the planner may ignore this advice.
Every arm, including the ToolGraph and flat-retrieval arms, exposes the full domain
tool catalog to the planner and performs no retrieval narrowing of the callable
action space, so differences between arms cannot be attributed to candidate-set
pruning.  The top-\(k\) block is therefore \emph{advice within} the prompt, not a
restriction of it --- which is why the flat-retrieval arm can score \emph{below}
Native (Section~\ref{sec:results}): it adds a block of text without removing any
option.  Native and ReAct receive the same catalog and policy but no graph features.
\textbf{Reflexion} here is the single-revision variant: it receives one
feedback-and-revision opportunity, so at a single trial its post-episode
mechanism is close to inert.  Its self-reflection is written by the same verifier
model, with the same revision budget, as the ToolGraph arms' verbalizer, so the
Reflexion-versus-ToolGraph difference isolates graph-grounded feedback from
ungrounded self-reflection.  Candidate temperature is 0.7 for all sampling-based
methods, except where a backbone's API rejects the parameter, in which case the
provider default applies.

\paragraph{Evaluation.}
A single seed controls planner sampling, tie breaking, and environment randomness;
task IDs, prompts, model versions, graph hashes, and generated candidates are cached
so that estimator comparisons use paired inputs.  Success is the benchmark-native
binary reward, and the reported cells are point estimates without confidence
intervals: with one seed per cell we have no within-cell variance to report, and the
run-to-run spread of a single configuration in Table~\ref{tab:sweeps} is the only
empirical handle on noise we can offer.  Token counts include scoring and
verbalization prompts, and graph cost is reported both in total and divided by the
number of evaluated tasks to show the effect of reuse.  We do not report per-decision
latency or probe coverage.

\end{document}